\documentclass{article} %
\usepackage[final]{colm2026_conference} %
\usepackage{microtype}
\usepackage{booktabs,amsmath,amssymb}
\usepackage{url}
\usepackage{hyperref}
\definecolor{darkblue}{rgb}{0, 0, 0.5}
\hypersetup{colorlinks=true, citecolor=darkblue, linkcolor=darkblue, urlcolor=darkblue}
\usepackage{graphicx} %

\title{BudgetSchemaBench: A Budget-Swept Diagnostic for Schema Context in Text-to-SQL}
\author{Chen Shen \\ Megagon Labs \\ \texttt{chen\_s@megagon.ai}}
\begin{document}
\maketitle
\lhead{Accepted at the COLM 2026 Workshop on Context Beyond the Window}

\begin{abstract}
Data agents over structured sources must fit database schema into the model's context window. Large catalogs
can span many databases and thousands of columns, so cost constraints may require choosing between table
coverage and serialization detail well before the context window is full. We introduce
\textbf{BudgetSchemaBench}, an execution-grounded diagnostic for this setting. Its construction derives
relevance labels mechanically from gold SQL, without human- or LLM-authored ground truth. Using a pooled
80-database catalog, we sweep four schema-context budgets and compare three representations while keeping
each retriever's table ranking fixed. A source-namespace check rejects queries that obtain the correct
result from the wrong database. The evaluation covers three conditions: end-to-end retrieval;
frozen-gold, in which the required tables are guaranteed; and a probe that removes those tables. For the
primary solver with raw serialization, raising the budget from 2.5\% to 50\% of the catalog improves
execution accuracy on 1,279 held-out questions by 18 percentage points under lexical retrieval but only
3 under dense retrieval; the dense retriever already finds most required tables at the smallest budget.
When the required tables are removed, 94.6\% of correct predictions name one of them exactly, consistent
with reconstruction of absent schema from parametric knowledge. For the two main solvers in the
frozen-gold condition, the three representations differ by at most 2 percentage points, and the widest
paired 95\% confidence interval bounds the difference within $\pm4$ points. We observe the same
qualitative patterns with one reasoning model from a different family. When retrieval is coverage-limited,
execution accuracy is more sensitive to the schema budget than to the tested serializations. The
diagnostic and the code used to construct and evaluate it are publicly available.
\end{abstract}

\section{Introduction}
Agents over structured data must place database schemas, catalogs, and metadata into the model's
context. At realistic scale, the relevant schema can exceed any practical schema-context
budget: an enterprise catalog spans many databases and thousands of columns, so the agent must
select and serialize schema under a token
budget. This is a distinct ``context beyond the window'' regime (typed, hierarchical, and
abstractable), yet the community lacks a shared way to measure it. Schema linking, union retrieval,
and prompt compression each address part of this problem, but no public diagnostic fixes a large
catalog, sweeps an explicit schema-context budget $K$, and reports downstream execution accuracy
while isolating the schema representation.

We contribute \textbf{BudgetSchemaBench}, a reusable, execution-grounded diagnostic for schema context
under a budget, together with the methodology that builds it. Our contributions are:
\textbf{(C1)}~an execution-grounded diagnostic that sweeps the schema-context budget $K$ and isolates
the effect of schema representation, with derivation-first labels, a source-namespace execution
contract, and a remove-the-needed-table probe; \textbf{(C2)}~a curation method that derives relevance
labels mechanically from gold SQL without human or LLM annotation and applies explicit
construction-validation gates; and \textbf{(C3)}~an evaluation on a held-out complement that exercises
the protocol end to end. We claim a diagnostic and a methodology. The tested retrievers and schema
representations are controlled experimental factors, and we make no claim that any one of them is
best. Section~\ref{sec:val} reports the full, independently audited held-out-complement evaluation:
$N{=}1{,}279$ for the end-to-end and frozen-gold conditions, with the remove-the-needed-table probe
evaluated separately. These results show how the diagnostic responds to changes in schema budget,
table availability, and representation.

\section{The missing evaluation object}
Three strands of prior work address parts of this evaluation problem: large-catalog retrieval,
controlled context-length evaluation, and schema representation (Table~\ref{tab:related}).
\citet{crush4sql} establish union-catalog construction with gold-SQL-derived labels on the SpiderUnion
and BirdUnion datasets and report retrieval recall under a retrieval budget. For open-domain
text-to-SQL, \citet{murre} extend table retrieval with a multi-hop method. These works
establish the large-catalog retrieval setting, but not execution-accuracy$(K)$ under representation
isolation, a source-namespace validity contract, or a remove-needed-table probe.
\citet{ruler} sweep context length as a controlled axis, but evaluate \emph{prose} rather than
relational catalogs with executable gold.

Large-catalog schema-linking systems \citep{linkalign,rasl,credsql} fix a large catalog and report EX,
but evaluate retriever methods rather than a controlled schema-budget axis with representation isolation.
A parallel line makes schema linking itself \emph{token-efficient}, pruning or re-ranking schema to cut
prompt cost while preserving recall (\citet{chess} progressively selects columns; \citet{rslsql}
bidirectional schema linking; \citet{extractivesl} an extractive selector; \citet{autolink} schema
exploration for scalable linking). That line studies the same recall-vs-token trade-off, but at
method-specific operating points rather than through a swept budget axis with execution-grounded
representation isolation. Closest in spirit, \citet{kasla} cast schema linking as a \emph{knapsack}
selection under a capacity tolerance, a budgeted method rather than a swept evaluation axis.

Schema-representation studies vary format inside full pipelines \citep{mschema,macsql} or use
representation diversity as an ensembling lever \citep{nrep}; they do not jointly control budget, isolate
representation, and report execution-grounded accuracy. Large-catalog realism benchmarks
\citep{spider2,beaver}, and concurrent enterprise-scale ones \citep{bothends}, raise schema difficulty,
scale, and cost, but do not sweep a controlled budget axis or isolate representation. Finally,
derivation-first labels inherit any noise in the source gold SQL: \citet{jin} report $52.8\%$ annotation
errors on BIRD's Mini-Dev subset, corroborating an earlier BIRD-Bench noise examination by
\citet{wretbladnoise}. We disclose this as a limitation; our parser-coverage and source-namespace gates
catch parser/namespace failures, but do not correct semantically wrong gold SQL.

Synthetic column-description work \citep{wretblad} isolates the effect of enrichment and finds that it
improves EX, but studies small schemas where the full catalog fits in the prompt and no budget binds.
We compare schema representations under a budget that prevents the full catalog from fitting.
\citet{maamari} show that strong models tolerate irrelevant schema under perfect recall. In our setting,
the schema budget binds, so some schema must be omitted and perfect recall cannot be guaranteed.
\citet{chung} examine whether long context windows remove the need for schema selection by placing more
schema in the prompt. They report that accuracy remains robust to irrelevant schema, although including
more context increases latency and cost even when the catalog fits. Our evaluation instead covers the
regime in which the full catalog cannot be included under the schema budget. In that regime, coverage
becomes the binding variable for evaluating schema representations.

\begin{table}[t]\centering\small
\begin{tabular}{lccc}\toprule
 & large catalog & budget-swept axis & repr.-isolated EX \\\midrule
CRUSH4SQL & \checkmark & recall@$K$ & --- \\
MURRE & \checkmark & EX, unswept & --- \\
RULER & prose & \checkmark & prose \\
Death of Schema Linking? & --- & --- & noise, not budget \\
Synthetic descriptions (Wretblad) & --- & --- & \checkmark (fits-window) \\
LinkAlign / RASL / CRED-SQL & \checkmark & --- & retriever, not repr. \\
\textbf{BudgetSchemaBench} & \checkmark & \checkmark & \checkmark \\\bottomrule
\end{tabular}
\caption{Where prior work stops. To our knowledge, BudgetSchemaBench is the first to combine a fixed
large catalog, the budget as a controlled swept axis, and execution-grounded representation isolation
(plus a source-namespace validity contract and a remove-needed-table probe).}\label{tab:related}
\end{table}

\section{BudgetSchemaBench}
\textbf{Released artifact.} We release the diagnostic under a BSD 3-Clause license as separately
addressable components, together with a content-hashed manifest, a pinned dependency lockfile, and a
one-command reproduction procedure using a versioned BIRD~\citep{bird} snapshot:
\begin{itemize}\itemsep1pt
\item versioned union-catalog build scripts and a pinned serializer $+$ tokenizer (\texttt{cl100k})%
\footnote{Budgets are measured in \texttt{cl100k} as a fixed, model-independent ruler; the catalog
differs by only $0.2\%$ under the native \texttt{o200k} tokenizer, so the budget axis is
tokenizer-insensitive across the OpenAI family. The cross-family \texttt{DeepSeek-V4-Flash} run
(\S\ref{sec:val}) is budgeted in the same \texttt{cl100k} units for comparability, since DeepSeek's
native tokenizer differs.};
\item namespace-rewritten gold SQL and the role-tagged derivation-first label schema;
\item the source-namespace evaluator that enforces the validity contract;
\item the \texttt{raw}/\texttt{enriched}/\texttt{hybrid} serializers and the fixed $K$ configurations;
\item the generation prompt template and the per-cell tokens-consumed logs;
\item the four construction-validation reports and the cross-database collision graph
(Table~\ref{tab:catalog}).
\end{itemize}
Every component above is available in the public release.\footnote{\url{https://github.com/megagonlabs/budget-schema}}
The construction-validation gates below run over the released artifact.

\textbf{Catalog and budget axis.} We pool databases into one canonical namespace (each table is
\texttt{<db>\_\_<table>}; Table~\ref{tab:catalog}) and sweep $K$ over four budgets, set as
absolute token caps identical across representations, equal to 2.5/10/25/50\% of the catalog
under the rich serialization ($127{,}560$ tokens; the \texttt{raw} catalog is $97{,}144$). The budget
$K$ caps the serialized schema payload only; the question, evidence, fixed instructions, and answer
scaffold sit outside $K$, so the sweep varies schema coverage with all else held constant. Tables are
packed greedily in retriever-rank order, with ties following the retriever's order. A table is included
whole if its serialization fits the remaining budget and skipped otherwise; tables are neither
truncated nor column-pruned, so the scan continues to smaller, lower-ranked tables. Because the same
$K$ cap applies across representations, richer serializations fit fewer tables;
Section~\ref{sec:val} examines this net-utility-under-budget effect.

The catalog is the BIRD dev$+$train union (80 databases). The $1{,}534$ evaluation questions come from
BIRD-dev's 11 databases, so the other 69 train databases serve purely as distractors. Of these
questions, $1{,}532$ are executable after excluding two with errors in the gold SQL. We split the
$1{,}534$ into a $255$-question development set (used only for calibration and the blind error study,
$n{=}179$ of these) and a disjoint $1{,}279$-question \emph{held-out complement} evaluated in
Section~\ref{sec:val}. The two questions with errors in the gold SQL are excluded from the
gold-invariance gate as benchmark errors.

The catalog need not exceed the physical context window for the diagnostic to be valid: $K$ is the
controlled budget, and the full schema never fits any $K$;
at enterprise scale a budget binds through retrieval cost and coverage well before any fixed window fills.
Pooling induces cross-database name collisions ($50$ colliding base-name groups spanning $146$ tables, $24.5\%$ of all tables), and the
\texttt{<db>} namespace prefix is a known signal. \emph{The main-text results use the prefixed
catalog}. Appendix~\ref{app:unprefixed} reports results for an unprefixed catalog, which removes the
namespace signal and provides a cleaner test of external validity because the prefix can make the
union easier than an unprefixed enterprise catalog. Under this condition, EX still increases with the
budget under lexical retrieval (large-minus-tiny ${+}0.157$, with a confidence interval that excludes
zero), whereas dense retrieval remains near saturation (${+}0.025$); the prefix contributes a small,
uniform $\sim$0.03--0.05 EX. The construction is limited to this BIRD snapshot, whose catalog size was
chosen as a tractable pilot; evaluation on larger catalogs remains future work.

\begin{table}[t]\centering\small
\begin{tabular}{lr}\toprule
databases (BIRD snapshot) & $80$ \\
tables & $597$ \\
tokens, full catalog (\texttt{cl100k}; \texttt{raw} $97{,}144$) & $127{,}560$ \\
colliding base-name groups / tables in collision & $50\,/\,146$ \,($24.5\%$) \\
budgets $K$ (\% of catalog) & $2.5\,/\,10\,/\,25\,/\,50$ \\
\quad{\footnotesize as \texttt{cl100k} tokens, to the nearest $100$} & $3200\,/\,12800\,/\,31900\,/\,63800$ \\\midrule
parser coverage & $100\%$ \;($1534/1534$) \\
gold-invariance (executable gold reproduces source) & $100\%$ \;{\footnotesize($1532/1532$; $2$ gold-bug Qs)} \\
recall sanity check (lexical; $n{=}400$ build sample), $K_\mathrm{tiny}\!\to\!K_\mathrm{large}$ & $0.75\!\to\!1.0$ \\
\bottomrule
\end{tabular}
\caption{\textbf{Union catalog at a glance} (the pilot instantiation). The full catalog never fits any
budget $K$; the namespace prefix and the cross-database collisions are reported because they make a
pooled catalog easier or harder than a real enterprise one. The bottom rows are the construction-%
validation gates (\S\ref{sec:val}), computed over the released artifact.}\label{tab:catalog}
\end{table}

\textbf{Retrieval.} Two representation-independent rankers score tables against the question using
table and column names only: a lexical name-bag-of-words ranker and a dense ranker based on OpenAI
\texttt{text-embedding-3-small}, which embeds the same table$+$column name string and ranks by cosine
similarity without fine-tuning or BIRD-specific training. Because both rankers use names only, their
table rankings are identical across representations; for packing under $K$, only the per-table payload
size changes. The design therefore isolates the effect of representation on generation; by
construction, richer descriptions cannot improve retrieval scores or rankings, so enriched
representations receive no retrieval-side benefit. End-to-end retrieval quality and
representation-aware retrieval remain outside the scope of this study and are discussed in the
limitations.

\textbf{Derivation-first labels, with no human or LLM ground truth.} We parse each question's gold SQL
to identify its relevant tables and columns, following the established schema-linking-evaluation
convention, and assign each element a role (selected/join/condition/order/group). A parser-coverage gate
excludes questions the parser cannot handle; on our substrate, it excludes $0\%$ after a CTE-aware
extractor fix. The resulting relevance labels are mechanically derived rather than authored by humans
or models. The enriched serializer uses the benchmark's human-written column descriptions only as
optional input features; neither the serialized schema content nor the relevance labels contain
model-generated text or annotations.

\textbf{Source-namespace execution: the construct-validity contract.} This contract is what makes
union-catalog EX measurable. Pooling many databases makes result-equivalent but wrong-source answers
easy to produce because the same value can recur across tables. Gold SQL is rewritten into the union
namespace and must reproduce its source-DB result; predicted SQL executes unchanged; correctness
requires both result-equivalence and source-namespace \emph{lineage}, so a result-equivalent but
wrong-source query is counted as an error. The contract is a narrow, low-cost validity check rather
than a source of metric noise. In the held-out evaluation, it changes the classification of only
$0.3\%$ of otherwise result-correct end-to-end predictions ($2$ of $797$ sampled predictions). Wrong-source
predictions are rare in this sample; by construction, the lineage requirement rejects such predictions.
Lineage is decided by static parse
(\texttt{sqlglot}): every base table the predicted SQL reads (across joins, sub-queries, and set
operations, with CTE-local names exempted) must carry the gold database's \texttt{<db>} prefix, and a
prediction that reads no source table is rejected. Result equivalence follows BIRD's official
metric (a set comparison of result rows~\citep{bird}), and our primary metric, \textbf{EX}, adds
the lineage requirement. We report a three-variant \emph{bracket} in which every variant requires
source-namespace lineage: a \emph{strict} variant (order- and multiplicity-sensitive), the BIRD
set-based \textbf{EX} (primary), and a \emph{relaxed} variant tolerant of output-column projection.
Every result below holds across all three variants; the bracket has a median width of $\sim$5 points,
is usually less than 8 points wide, and has a maximum width of $\sim$10 points.

A blind error study on the model-calibration set ($n{=}179$ questions, three solvers) categorized the
observed failure modes. Roughly a third of the errors pooled across the three solvers ($\approx$65) arose from projection or
format-convention mismatches or from errors in the gold SQL, all orthogonal to schema understanding.
These errors affect all representations equally but widen CIs and depress absolute EX. The relaxed
variant can credit projection-related mismatches among them.

\textbf{Three controlled regimes.} (i)~\emph{End-to-end}: rank the catalog, fill to $K$, let the
budget determine retrieval coverage (the budget-axis test). (ii)~\emph{Frozen-gold}: guarantee the
gold tables, then fill the remaining budget. Table serialization varies by design to isolate its effect
on generation. Under a fixed budget, however, a longer serialization admits fewer non-gold
distractors. This contrast therefore measures net utility under budget rather than a pure serialization
swap. A fixed-table-set control remains future work. (iii)~\emph{Remove-the-needed-table probe}:
exclude the gold tables from the catalog to assess sensitivity to missing needed schema.

\textbf{Representations and reporting.} \texttt{raw} (DDL + sample rows), \texttt{enriched}
(\,$+$\,the benchmark's human-written column descriptions), \texttt{hybrid} (budget-aware mix). The protocol
\emph{records}, per cell, the EX bracket$(K)$ (strict / set / relaxed), table recall$(K)$,
tokens-consumed as a proxy for the cost axis, and collision-aware strata (all / unambiguous / ambiguous). The
held-out-complement evaluation below \emph{reports} the EX bracket and table recall$(K)$ with bootstrap
CIs over questions (and a database-clustered bootstrap); role-stratified column recall and a
collision-strata breakdown remain future work (Section~5).

\section{Construction validation and held-out-complement evaluation}\label{sec:val}
\textbf{Construction validation.} The catalog exceeds $K_\mathrm{large}$, so no tested budget
admits the full schema. The gold-SQL label parser excludes $0\%$ of questions; every executable
rewritten gold query reproduces its source-database result (gold invariance $1532/1532$ after the
CTE-aware extractor fix; two questions with errors in the gold SQL, $<\!0.2\%$, are excluded as
benchmark errors). In a build-stage recall check over $400$ sampled questions, lexical recall rises from $0.75$ to
$1.0$ across the four budgets, so the budget excludes needed tables at $K_\mathrm{tiny}$. The source-namespace tests reject result-equivalent
wrong-source answers. Together, these checks establish the construct validity required for the
held-out evaluation.

\textbf{Held-out-complement evaluation (independently audited).} We evaluate on the full held-out
complement, disjoint from development: $N{=}1{,}279$ exactly for end-to-end and frozen-gold; the
remove-the-needed-table probe over its $1{,}236$ predictions counted as correct, discussed below; and
a frontier-model slice at $N{=}30$. The primary solver is a GPT-5-class reasoning model,
\texttt{gpt-5.4-mini}, run with low reasoning effort and sampling temperature $1$.\footnote{The
primary evaluation uses a fixed dated snapshot and one decode per question; run IDs and seeds are
released with the artifact.} For comparison, we also evaluate a non-reasoning model
(\texttt{gpt-4o-mini}, greedy / temperature $0$), a frontier slice (\texttt{gpt-5.5}, $N{=}30$,
frozen-gold only), and an off-family reasoning model (\texttt{DeepSeek-V4-Flash};
\S\ref{sec:val} ``cross-family replication''). An independent result-stage audit was conducted by a
model family different from the one that produced the analysis.
Bootstrap $95\%$ CIs accompany every EX estimate and every paired contrast; table-recall values are
point estimates. Table~\ref{tab:budget} (the budget axis) and Table~\ref{tab:repr} (the frozen-gold
representation contrast) give the per-cell grid for those two conditions.

\begin{figure}[t]\centering
\includegraphics[width=0.8\linewidth]{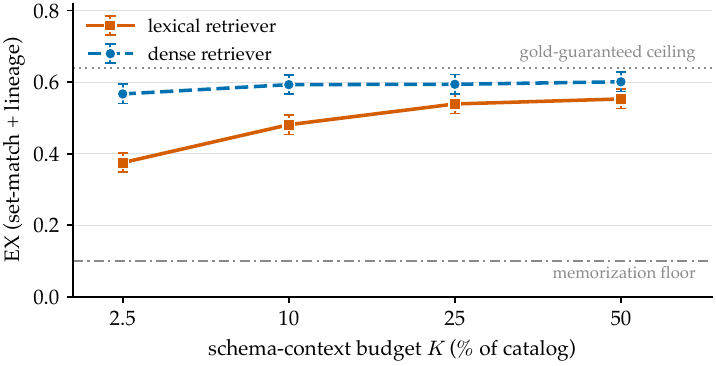}
\caption{\textbf{Execution accuracy increases with the schema budget under coverage-limited
retrieval.} End-to-end EX$(K)$ on the full held-out complement ($N{=}1{,}279$, \texttt{raw},
\texttt{gpt-5.4-mini}; bootstrap $95\%$ CIs; the increase holds across the strict/relaxed bracket).
The lexical retriever (solid) rises with
$K$, tracking table-recall $0.60\!\to\!0.90$; the dense retriever (dashed) keeps recall high
($0.92\!\to\!1.0$) even at the tightest budget and stays near the gold-guaranteed ceiling, flattening the
curve. The dotted reference line is EX with the gold tables guaranteed; the dash-dotted line is the
lexical schema-memorization floor with the gold tables removed ($0.10$ at the $25\%$ budget, a likely
leakage level). The dense floor is higher and is not drawn ($0.18$ at the same budget); across $K$ the
two span $0.04$--$0.19$. The conditions differ in retrieval mechanism, so
this is a sensitivity contrast, not a causal decomposition.}\label{fig:bracket}
\end{figure}

\textbf{Execution accuracy increases with the schema budget under coverage-limited retrieval.}
Under lexical retrieval, EX rises with $K$ ($0.38\!\to\!0.55$, $N{=}1{,}279$), tracking table recall
($0.60\!\to\!0.90$, Figure~\ref{fig:bracket}). For both models, the paired large-minus-tiny gains
have $95\%$ confidence intervals that exclude zero
($\Delta{=}{+}0.18\,[0.15,0.20]$ and ${+}0.09\,[0.07,0.12]$); the corresponding intervals also
exclude zero under a database-clustered bootstrap over the 11 source databases. The increase holds
across the bracket (strict $0.35\!\to\!0.52$, relaxed $0.39\!\to\!0.57$). Increasing the
schema-context budget raises EX when retrieval is coverage-limited; when a strong retriever already
finds the gold tables, a larger budget provides little additional improvement. EX also increases
with $K$ under lexical retrieval on the unprefixed catalog (Appendix~\ref{app:unprefixed}), which
removes the \texttt{<db>} source signal.

\textbf{Execution accuracy varies with gold-table availability.} EX is lowest when the gold tables
are removed, higher under end-to-end retrieval, and highest when the gold tables are guaranteed
(Figure~\ref{fig:bracket}). With the gold tables absent from context, the solver still answers
$4$--$19\%$ of questions correctly (lexical EX $0.04$--$0.10$ and dense EX $0.16$--$0.19$ across
$K$) and in some cases refers to removed tables by name. We interpret this nonzero performance as a
likely leakage level and call it the \emph{schema-memorization floor}. A per-question audit of the
remove-gold results supports exact-name schema reconstruction. Of the $1{,}236$ predictions counted
as correct (\texttt{gpt-5.4-mini}; $365$ unique questions, each counted in one or more
budget/retriever cells), \textbf{94.6\%} ($1{,}169/1{,}236$) refer to a removed gold table by its
exact name, consistent with reconstruction of absent schema from parametric (weight-time) memory.
The exact-name flag was computed mechanically over the full set, and manual inspection confirmed all
15 sampled flags (15/15). The remaining $5.4\%$ is not adjudicated by this audit; these are mostly questions answerable from
data values that remain available, a weaker leakage channel. Because the conditions differ in
retrieval mechanism, this comparison is a perturbation/sensitivity contrast rather than a causal
decomposition. Exact-name
overlap is consistent with weight-time memorization but does not by itself separate it from generic
schema-name priors; an identifier-randomization control, and auditing all solvers with a value-leakage
control, are future work (Section~5).

\textbf{Representation differences are small and bounded in the frozen-gold condition.} In this
retrieval-isolated setup, the tested serializations show no practically large generation-side effect.
Across the full frozen-gold evaluation ($N{=}1{,}279$; Table~\ref{tab:repr}), paired
\texttt{raw}/\texttt{enriched}/\texttt{hybrid} differences are small
(max $|\Delta|{=}0.02$), with no contrast surviving Holm correction
(every $p_\mathrm{Holm}{\ge}0.27$); the widest paired $95\%$ CI bounds any representation effect to
within $\pm0.04$ EX. We report this measured bound rather than a formal $\delta$-equivalence test;
the frontier slice (\texttt{gpt-5.5}, $N{=}30$) is wider (point $|\Delta|$ up to $0.07$, CIs to
$\pm0.20$) and illustrative only. We make no claim that any representation is superior or
equivalent. Under coverage-limited retrieval, execution accuracy is more sensitive to the schema
budget than to the tested serializations. Consistent with this, a stronger (dense) retriever keeps
end-to-end recall high ($0.92\!\to\!1.0$) even at the
tightest budget, which largely saturates the $K$ curve (Table~\ref{tab:budget}).

\begin{table}[t]\centering\small\setlength{\tabcolsep}{4pt}
\begin{tabular}{l cc cc cc}\toprule
 & \multicolumn{2}{c}{\texttt{gpt-5.4-mini} EX} & \multicolumn{2}{c}{\texttt{gpt-4o-mini} EX}
 & \multicolumn{2}{c}{recall} \\
\cmidrule(lr){2-3}\cmidrule(lr){4-5}\cmidrule(lr){6-7}
budget $K$ & lexical & dense & lexical & dense & lex. & dense \\\midrule
2.5\% & $0.38${\scriptsize$[.35,.40]$} & $0.57${\scriptsize$[.54,.59]$} & $0.30${\scriptsize$[.28,.33]$} & $0.48${\scriptsize$[.45,.51]$} & $0.60$ & $0.92$ \\
10\%  & $0.48${\scriptsize$[.45,.51]$} & $0.59${\scriptsize$[.57,.62]$} & $0.37${\scriptsize$[.34,.40]$} & $0.49${\scriptsize$[.46,.52]$} & $0.81$ & $0.98$ \\
25\%  & $0.54${\scriptsize$[.51,.57]$} & $0.59${\scriptsize$[.57,.62]$} & $0.40${\scriptsize$[.37,.43]$} & $0.48${\scriptsize$[.45,.51]$} & $0.89$ & $1.00$ \\
50\%  & $0.55${\scriptsize$[.53,.58]$} & $0.60${\scriptsize$[.57,.63]$} & $0.39${\scriptsize$[.37,.42]$} & $0.47${\scriptsize$[.44,.50]$} & $0.90$ & $1.00$ \\
\bottomrule
\end{tabular}
\caption{\textbf{End-to-end budget sweep} (full held-out complement, $N{=}1{,}279$, \texttt{raw} serialization):
EX$(K)$ (BIRD set-match $+$ lineage) with bootstrap $95\%$ CIs, plus table-recall$(K)$, for two solvers under
lexical vs.\ dense retrieval. Under lexical retrieval, the large-minus-tiny gains have $95\%$
confidence intervals that exclude zero ($\Delta{=}{+}0.18\,[0.15,0.20]$ for
\texttt{gpt-5.4-mini}, ${+}0.09\,[0.07,0.12]$ for \texttt{gpt-4o-mini}; both $p{<}10^{-3}$). The
corresponding gains are much smaller under dense retrieval ($\Delta{=}{+}0.03\,[0.02,0.05]$ for
\texttt{gpt-5.4-mini}; ${-}0.01\,[{-}0.03,0.01]$, not separable, for
\texttt{gpt-4o-mini}).}
\label{tab:budget}
\end{table}

\begin{table}[t]\centering\small\setlength{\tabcolsep}{5pt}
\begin{tabular}{l ccc}\toprule
budget $K$ & \texttt{raw} & \texttt{enriched} & \texttt{hybrid} \\\midrule
\multicolumn{4}{l}{\emph{\texttt{gpt-5.4-mini}}}\\
2.5\% & $0.65${\scriptsize$[.62,.67]$} & $0.65${\scriptsize$[.62,.68]$} & $0.65${\scriptsize$[.63,.68]$} \\
10\%  & $0.64${\scriptsize$[.62,.67]$} & $0.66${\scriptsize$[.64,.69]$} & $0.65${\scriptsize$[.62,.67]$} \\
25\%  & $0.63${\scriptsize$[.60,.65]$} & $0.64${\scriptsize$[.62,.67]$} & $0.64${\scriptsize$[.61,.67]$} \\
50\%  & $0.63${\scriptsize$[.61,.66]$} & $0.64${\scriptsize$[.61,.67]$} & $0.63${\scriptsize$[.60,.66]$} \\
\midrule
\multicolumn{4}{l}{\emph{\texttt{gpt-4o-mini}}}\\
2.5\% & $0.58${\scriptsize$[.55,.60]$} & $0.59${\scriptsize$[.56,.62]$} & $0.59${\scriptsize$[.56,.62]$} \\
10\%  & $0.56${\scriptsize$[.53,.59]$} & $0.58${\scriptsize$[.56,.61]$} & $0.57${\scriptsize$[.55,.60]$} \\
25\%  & $0.56${\scriptsize$[.53,.58]$} & $0.57${\scriptsize$[.54,.60]$} & $0.57${\scriptsize$[.54,.59]$} \\
50\%  & $0.54${\scriptsize$[.51,.57]$} & $0.55${\scriptsize$[.52,.57]$} & $0.55${\scriptsize$[.52,.58]$} \\
\bottomrule
\end{tabular}
\caption{\textbf{Frozen-gold representation contrast} (full held-out complement, $N{=}1{,}279$): with the gold
tables guaranteed, only the per-table serialization varies. EX$(K)$ with bootstrap $95\%$ CIs
for \texttt{raw}, \texttt{enriched}, and \texttt{hybrid}. Across all $24$ paired contrasts (the two main solvers) the largest
point difference is $|\Delta|{=}0.02$, no contrast survives Holm correction (every $p_\mathrm{Holm}{\ge}0.27$), and the widest paired $95\%$ CI
reaches $\pm0.04$ EX: a bounded null at this scale, not an established equivalence (the frontier slice
\texttt{gpt-5.5}, $N{=}30$, is wider and illustrative only).}\label{tab:repr}
\end{table}

\textbf{Cross-family replication (a non-GPT reasoning solver).} To test whether these patterns are
specific to the GPT family, we ran the complete protocol once more with an off-family reasoning
model, \texttt{DeepSeek-V4-Flash}.\footnote{Served through Fireworks serverless; run IDs are released
with the artifact.} The provider maps the requested \texttt{low} reasoning-effort setting to
\emph{high}. The model is therefore reasoning-enabled but not effort-matched to
\texttt{gpt-5.4-mini}; this run provides one off-family replication rather than evidence of broad
cross-model generality. The off-family model reproduces all three qualitative patterns
(Table~\ref{tab:deepseek}): the lexical budget axis rises by the same margin
(large-minus-tiny $\Delta{=}{+}0.18$), while the dense axis stays nearly flat ($+0.06$); frozen
\texttt{raw}/\texttt{enriched}/\texttt{hybrid} agree within $|\Delta|{\le}0.05$; and the remove-gold
floor is low (lexical $0.01$--$0.05$, dense $0.05$--$0.13$ EX). This replication suggests that the
observed patterns are not specific to the GPT family.

\begin{table}[t]\centering\small\setlength{\tabcolsep}{6pt}
\begin{tabular}{lcc}\toprule
diagnostic pattern & DeepSeek-V4-Flash & (\texttt{gpt-5.4-mini}) \\\midrule
budget axis, lexical (large-minus-tiny $\Delta$) & $+0.18$ & $(+0.18)$ \\
budget axis, dense (large-minus-tiny $\Delta$)   & $+0.06$ & $(+0.03)$ \\
frozen representation (max $|\Delta|$)        & $0.05$  & $(0.02)$ \\
memorization floor, lexical (EX across $K$)   & $0.01$--$0.05$ & $(0.04$--$0.10)$ \\
memorization floor, dense (EX across $K$)     & $0.05$--$0.13$ & $(0.16$--$0.19)$ \\
\bottomrule
\end{tabular}
\caption{\textbf{Cross-family replication.} An off-family reasoning model
(\texttt{DeepSeek-V4-Flash}, reasoning-enabled) reproduces all three qualitative patterns on the
full held-out complement (EX $=$ BIRD set-match $+$ lineage); the \texttt{gpt-5.4-mini} value is shown
in brackets for reference. This is one off-family replication, not a broad-generality claim.}
\label{tab:deepseek}
\end{table}

\section{Limitations}
The main end-to-end and frozen-gold evaluations cover the full held-out complement
($N{=}1{,}279$), with a \texttt{gpt-5.5} frontier slice at $N{=}30$; the collision-strata
breakdown remains future work. The evaluated solvers include a reasoning GPT model, a
non-reasoning GPT model, and an off-family reasoning model
(\texttt{DeepSeek-V4-Flash}). The off-family model reproduces all three qualitative patterns,
and the \texttt{gpt-5.5} frontier slice is consistent with them. These results suggest that the
patterns are not specific to the GPT family; broader model and benchmark coverage remains
future work.

The remove-gold probe yields nonzero performance that we interpret as a likely leakage level
and call the schema-memorization floor. In the per-question audit, $94.6\%$ of predictions
counted as correct refer to removed gold tables by exact name, consistent with reconstruction
of absent schema from parametric memory; a small value-leakage residual remains. Extending
the audit to all solvers and adding a value-leakage control remain future work.

For the representation result, we report the measured paired-CI bound ($\pm0.04$ EX) rather
than a formal equivalence verdict from two one-sided tests (TOST). The preregistered
$\delta{=}0.03$ equivalence margin is not met at this $N$ because the widest paired interval
($\pm0.04$) exceeds it. Accordingly, we report a bounded null without claiming equivalence.

Two further reporting caveats apply. We evaluate the budget effect using both a question-level
and a database-clustered bootstrap. The two agree, but the database-clustered analysis includes
only 11 source databases. The primary solver is evaluated with a single decode at
temperature~$1$, so decode-sampling variance is not separately modelled.

The representation-independent rankers isolate the effect of representation on generation by
preventing representation content from affecting retrieval scores or rankings. The design
therefore does not evaluate representation-aware retrieval. The EX gain reported for synthetic descriptions
\citep{wretblad} is a generation-channel gain measured with the full schema in prompt and no budget
binding, so our bounded null holds only under this budgeted, retrieval-isolated setup and does not
settle enrichment's value where representation can also change what is retrieved. The derivation-first methodology (C2) transfers only to settings that already
provide gold SQL (real catalogs need query-log mining or labels); union catalogs introduce
cross-database collisions, which we measure and report; and gold-SQL labels inherit any non-minimal
joins in the source queries.

The off-the-shelf dense retriever already achieves table recall near $1.0$ at the tightest
budget, so this catalog may understate the difficulty of schema selection for strong
retrievers. Evaluation with semantically paraphrased queries or larger,
harder-to-disambiguate catalogs remains future work.

\section{Conclusion}
BudgetSchemaBench is a validated diagnostic for schema context under a budget, released with
an executable artifact and a curation methodology that derives relevance labels without
human- or LLM-authored ground truth. The full held-out-complement evaluation spans multiple
solvers and one off-family reasoning model and examines three questions: whether the budget
matters under coverage-limited retrieval, how much schema can be reconstructed when needed
tables are removed, and how much generation changes when retrieval is isolated. The evaluation
shows that execution accuracy increases with the schema budget under coverage-limited
retrieval, identifies a schema-memorization floor consistent with reconstruction from
parametric (weight-time) memory, and bounds differences among the tested representations in
the frozen-gold condition. The tested retrievers and schema representations are controlled
experimental factors; we make no claim that any retriever or representation is best. When
retrieval is coverage-limited, execution accuracy is more sensitive to the schema budget than
to the tested serializations. The released artifact lets benchmark builders and method
developers measure where a schema-context budget is spent.

\bibliographystyle{colm2026_conference}
\bibliography{refs}

\begin{thebibliography}{23}
\providecommand{\natexlab}[1]{#1}
\providecommand{\url}[1]{\texttt{#1}}
\expandafter\ifx\csname urlstyle\endcsname\relax
  \providecommand{\doi}[1]{doi: #1}\else
  \providecommand{\doi}{doi: \begingroup \urlstyle{rm}\Url}\fi

\bibitem[Cao et~al.(2024)Cao, Zheng, Fan, Zhang, Chen, and Bai]{rslsql}
Zhenbiao Cao, Yuanlei Zheng, Zhihao Fan, Xiaojin Zhang, Wei Chen, and Xiang
  Bai.
\newblock {RSL-SQL}: Robust schema linking in text-to-{SQL} generation.
\newblock \emph{arXiv preprint arXiv:2411.00073}, 2024.

\bibitem[Chen et~al.(2024)Chen, Yang, Li, Wenz, Zhang, Tatbul, Cafarella,
  Demiralp, and Stonebraker]{beaver}
Peter~Baile Chen, Devin Yang, Weiyue Li, Fabian Wenz, Yi~Zhang, Nesime Tatbul,
  Michael Cafarella, {\c{C}}a{\u{g}}atay Demiralp, and Michael Stonebraker.
\newblock {BEAVER}: An enterprise benchmark for text-to-sql.
\newblock \emph{arXiv preprint arXiv:2409.02038}, 2024.

\bibitem[Chung et~al.(2025)Chung, Kakkar, Gan, Milne, and {\"O}zcan]{chung}
Yeounoh Chung, Gaurav~T. Kakkar, Yu~Gan, Brenton Milne, and Fatma {\"O}zcan.
\newblock Is long context all you need? leveraging {LLM}'s extended context for
  {NL2SQL}.
\newblock \emph{Proceedings of the VLDB Endowment}, 18\penalty0 (8):\penalty0
  2735--2747, 2025.
\newblock \doi{10.14778/3742728.3742761}.

\bibitem[D{\"o}nder et~al.(2025)D{\"o}nder, Hommel, Wen-Yi, Mimno, and
  Jo]{nrep}
Yusuf~Denizay D{\"o}nder, Derek Hommel, Andrea~W. Wen-Yi, David Mimno, and Unso
  Eun~Seo Jo.
\newblock Cheaper, better, faster, stronger: Robust text-to-sql without
  chain-of-thought or fine-tuning.
\newblock \emph{arXiv preprint arXiv:2505.14174}, 2025.

\bibitem[Duan et~al.(2025)Duan, Wang, Liu, Zhu, Zhang, Han, Yan, and
  Peng]{credsql}
Shaoming Duan, Zirui Wang, Chuanyi Liu, Zhibin Zhu, Yuhao Zhang, Peiyi Han,
  Liang Yan, and Zewu Peng.
\newblock {CRED-SQL}: Enhancing real-world large scale database text-to-sql
  parsing through cluster retrieval and execution description.
\newblock In \emph{{ECAI} 2025: 28th European Conference on Artificial
  Intelligence}, volume 413 of \emph{Frontiers in Artificial Intelligence and
  Applications}, pp.\  4394--4401. IOS Press, 2025.
\newblock \doi{10.3233/FAIA251337}.

\bibitem[Eben et~al.(2025)Eben, Ahmad, and Lau]{rasl}
Jeffrey Eben, Aitzaz Ahmad, and Stephen Lau.
\newblock {RASL}: Retrieval augmented schema linking for massive database
  text-to-sql.
\newblock \emph{arXiv preprint arXiv:2507.23104}, 2025.

\bibitem[Eizaguirre et~al.(2026)Eizaguirre, Tissen, and
  S{\'a}nchez-Artigas]{bothends}
Germ{\'a}n~T. Eizaguirre, Lars Tissen, and Marc S{\'a}nchez-Artigas.
\newblock Both ends count! just how good are {LLM} agents at ``text-to-big
  {SQL}''?
\newblock In \emph{Proceedings of the Sixth European Workshop on Machine
  Learning and Systems (EuroMLSys)}, pp.\  333--345. ACM, 2026.
\newblock \doi{10.1145/3805621.3807640}.
\newblock URL \url{https://doi.org/10.1145/3805621.3807640}.

\bibitem[Gao et~al.(2024)Gao, Liu, Li, Shi, Zhu, Wang, Li, Li, Hong, Luo, Gao,
  Mou, and Li]{mschema}
Yingqi Gao, Yifu Liu, Xiaoxia Li, Xiaorong Shi, Yin Zhu, Yiming Wang, Shiqi Li,
  Wei Li, Yuntao Hong, Zhiling Luo, Jinyang Gao, Liyu Mou, and Yu~Li.
\newblock A preview of {XiYan-SQL}: A multi-generator ensemble framework for
  text-to-sql.
\newblock \emph{arXiv preprint arXiv:2411.08599}, 2024.

\bibitem[Glass et~al.(2025)Glass, Eyceoz, Subramanian, Rossiello, Vu, and
  Gliozzo]{extractivesl}
Michael Glass, Mustafa Eyceoz, Dharmashankar Subramanian, Gaetano Rossiello,
  Long Vu, and Alfio Gliozzo.
\newblock Extractive schema linking for text-to-{SQL}.
\newblock \emph{arXiv preprint arXiv:2501.17174}, 2025.

\bibitem[Hsieh et~al.(2024)Hsieh, Sun, Kriman, Acharya, Rekesh, Jia, Zhang, and
  Ginsburg]{ruler}
Cheng-Ping Hsieh, Simeng Sun, Samuel Kriman, Shantanu Acharya, Dima Rekesh, Fei
  Jia, Yang Zhang, and Boris Ginsburg.
\newblock {RULER}: What's the real context size of your long-context language
  models?
\newblock In \emph{Proceedings of the First Conference on Language Modeling
  (COLM)}, 2024.

\bibitem[Jin et~al.(2026)Jin, Choi, Zhu, and Kang]{jin}
Tengjun Jin, Yoojin Choi, Yuxuan Zhu, and Daniel Kang.
\newblock Pervasive annotation errors break text-to-sql benchmarks and
  leaderboards.
\newblock \emph{Proceedings of the VLDB Endowment}, 19\penalty0 (5):\penalty0
  931--944, 2026.
\newblock \doi{10.14778/3796195.3796206}.

\bibitem[Kothyari et~al.(2023)Kothyari, Dhingra, Sarawagi, and
  Chakrabarti]{crush4sql}
Mayank Kothyari, Dhruva Dhingra, Sunita Sarawagi, and Soumen Chakrabarti.
\newblock {CRUSH4SQL}: Collective retrieval using schema hallucination for
  {Text2SQL}.
\newblock In \emph{Proceedings of the 2023 Conference on Empirical Methods in
  Natural Language Processing (EMNLP)}, 2023.

\bibitem[Lei et~al.(2025)Lei, Chen, Ye, Cao, Shin, Su, Suo, Gao, Hu, Yin,
  Zhong, Xiong, Sun, Liu, Wang, and Yu]{spider2}
Fangyu Lei, Jixuan Chen, Yuxiao Ye, Ruisheng Cao, Dongchan Shin, Hongjin Su,
  Zhaoqing Suo, Hongcheng Gao, Wenjing Hu, Pengcheng Yin, Victor Zhong, Caiming
  Xiong, Ruoxi Sun, Qian Liu, Sida Wang, and Tao Yu.
\newblock Spider 2.0: Evaluating language models on real-world enterprise
  text-to-sql workflows.
\newblock In \emph{Proceedings of the 13th International Conference on Learning
  Representations (ICLR)}, 2025.

\bibitem[Li et~al.(2023)Li, Hui, Qu, Yang, Li, Li, Wang, Qin, Geng, Huo, Zhou,
  Ma, Li, Chang, Huang, Cheng, and Li]{bird}
Jinyang Li, Binyuan Hui, Ge~Qu, Jiaxi Yang, Binhua Li, Bowen Li, Bailin Wang,
  Bowen Qin, Ruiying Geng, Nan Huo, Xuanhe Zhou, Chenhao Ma, Guoliang Li,
  Kevin~C.C. Chang, Fei Huang, Reynold Cheng, and Yongbin Li.
\newblock Can {LLM} already serve as a database interface? a {BIg} bench for
  large-scale database grounded text-to-sqls.
\newblock In \emph{Advances in Neural Information Processing Systems
  (NeurIPS)}, 2023.

\bibitem[Maamari et~al.(2024)Maamari, Abubaker, Jaroslawicz, and
  Mhedhbi]{maamari}
Karime Maamari, Fadhil Abubaker, Daniel Jaroslawicz, and Amine Mhedhbi.
\newblock The death of schema linking? text-to-sql in the age of well-reasoned
  language models.
\newblock \emph{arXiv preprint arXiv:2408.07702}, 2024.

\bibitem[Talaei et~al.(2024)Talaei, Pourreza, Chang, Mirhoseini, and
  Saberi]{chess}
Shayan Talaei, Mohammadreza Pourreza, Yu-Chen Chang, Azalia Mirhoseini, and
  Amin Saberi.
\newblock {CHESS}: Contextual harnessing for efficient {SQL} synthesis.
\newblock \emph{arXiv preprint arXiv:2405.16755}, 2024.

\bibitem[Wang et~al.(2025{\natexlab{a}})Wang, Ren, Yang, Liang, Bai, Chai, Yan,
  Zhang, Yin, Sun, and Li]{macsql}
Bing Wang, Changyu Ren, Jian Yang, Xinnian Liang, Jiaqi Bai, LinZheng Chai,
  Zhao Yan, Qian-Wen Zhang, Di~Yin, Xing Sun, and Zhoujun Li.
\newblock {MAC-SQL}: A multi-agent collaborative framework for text-to-sql.
\newblock In \emph{Proceedings of the 31st International Conference on
  Computational Linguistics (COLING)}, 2025{\natexlab{a}}.

\bibitem[Wang et~al.(2025{\natexlab{b}})Wang, Liu, and Yang]{linkalign}
Yihan Wang, Peiyu Liu, and Xin Yang.
\newblock {LinkAlign}: Scalable schema linking for real-world large-scale
  multi-database text-to-sql.
\newblock In \emph{Proceedings of the 2025 Conference on Empirical Methods in
  Natural Language Processing (EMNLP)}, pp.\  977--991. Association for
  Computational Linguistics, 2025{\natexlab{b}}.
\newblock \doi{10.18653/v1/2025.emnlp-main.51}.
\newblock URL \url{https://aclanthology.org/2025.emnlp-main.51/}.

\bibitem[Wang et~al.(2026)Wang, Zheng, Cao, Zhang, Wei, Fu, Luo, Chen, and
  Bai]{autolink}
Ziyang Wang, Yuanlei Zheng, Zhenbiao Cao, Xiaojin Zhang, Zhongyu Wei, Pei Fu,
  Zhenbo Luo, Wei Chen, and Xiang Bai.
\newblock {AutoLink}: Autonomous schema exploration and expansion for scalable
  schema linking in text-to-{SQL} at scale.
\newblock \emph{Proceedings of the {AAAI} Conference on Artificial
  Intelligence}, 40\penalty0 (40):\penalty0 33809--33817, 2026.
\newblock \doi{10.1609/aaai.v40i40.40672}.

\bibitem[Wretblad et~al.(2024{\natexlab{a}})Wretblad, Holmstr{\"o}m, Larsson,
  Wiks{\"a}ter, S{\"o}derlund, {\"O}hman, Pont{\'e}n, Forsberg, S{\"o}rme, and
  Heintz]{wretblad}
Niklas Wretblad, Oskar Holmstr{\"o}m, Erik Larsson, Axel Wiks{\"a}ter, Oscar
  S{\"o}derlund, Hjalmar {\"O}hman, Ture Pont{\'e}n, Martin Forsberg, Martin
  S{\"o}rme, and Fredrik Heintz.
\newblock Synthetic sql column descriptions and their impact on text-to-sql
  performance.
\newblock \emph{arXiv preprint arXiv:2408.04691}, 2024{\natexlab{a}}.

\bibitem[Wretblad et~al.(2024{\natexlab{b}})Wretblad, Riseby, Biswas, Ahmadi,
  and Holmstr{\"o}m]{wretbladnoise}
Niklas Wretblad, Fredrik~Gordh Riseby, Rahul Biswas, Amin Ahmadi, and Oskar
  Holmstr{\"o}m.
\newblock Understanding the effects of noise in text-to-{SQL}: An examination
  of the {BIRD}-bench benchmark.
\newblock In \emph{Proceedings of the 62nd Annual Meeting of the Association
  for Computational Linguistics (ACL), Short Papers}, 2024{\natexlab{b}}.

\bibitem[Yuan et~al.(2026)Yuan, Chen, Hong, Zhang, Huang, Li, and Huang]{kasla}
Zheng Yuan, Hao Chen, Zijin Hong, Qinggang Zhang, Feiran Huang, Qing Li, and
  Xiao Huang.
\newblock Knapsack optimization-based schema linking for {LLM}-based
  text-to-{SQL} generation.
\newblock In \emph{2026 {IEEE} 42nd International Conference on Data
  Engineering (ICDE)}, pp.\  531--544. IEEE, 2026.
\newblock \doi{10.1109/ICDE65706.2026.00046}.

\bibitem[Zhang et~al.(2025)Zhang, Wang, Dou, Zhu, and Che]{murre}
Xuanliang Zhang, Dingzirui Wang, Longxu Dou, Qingfu Zhu, and Wanxiang Che.
\newblock {MURRE}: Multi-hop table retrieval with removal for open-domain
  text-to-sql.
\newblock In \emph{Proceedings of the 31st International Conference on
  Computational Linguistics (COLING)}, 2025.

\end{thebibliography}

\appendix
\section{Unprefixed-catalog robustness}\label{app:unprefixed}
The main text uses the prefixed catalog, where each table is named \texttt{<db>\_\_<table>} and the
\texttt{<db>} prefix is a known source-database signal. To test whether the budget effect depends on
that signal, we repeat the end-to-end budget sweep (\texttt{raw}, \texttt{gpt-5.4-mini}, full
$N{=}1{,}279$, lexical and dense) over an \emph{unprefixed} catalog: each table is shown under its
bare table name, with the $146$ colliding tables ($24.5\%$) given a source-agnostic numeric suffix, so the model
sees no source-database signal. Retrieval, table-recall, the source-namespace lineage
contract, and EX all remain in the union namespace: only the prompt names change, and predictions are
translated back to the union namespace before the unchanged scorer. Table~\ref{tab:unprefixed} compares the
two catalog variants.

\textbf{The budget effect persists without namespace prefixes.} Under lexical retrieval, EX
still increases with $K$, and the $95\%$ confidence interval for the paired large-minus-tiny
gain excludes zero (unprefixed ${+}0.157\,[0.132,0.182]$ vs.\ prefixed ${+}0.178$). Under
dense retrieval, EX remains near saturation (${+}0.025\,[0.003,0.046]$ vs.\
${+}0.034$); table recall is identical by construction. The prefix contributes a small,
uniform absolute increase ($\sim$0.03--0.05 EX across all cells), consistent with providing
one additional disambiguation cue. The budget effect therefore does not depend on the
namespace prefix.

\begin{table}[!h]\centering\small\setlength{\tabcolsep}{5pt}
\begin{tabular}{l cc cc}\toprule
 & \multicolumn{2}{c}{lexical EX} & \multicolumn{2}{c}{dense EX} \\
\cmidrule(lr){2-3}\cmidrule(lr){4-5}
budget $K$ & unprefixed & prefixed & unprefixed & prefixed \\\midrule
2.5\% & $0.348${\scriptsize$[.32,.37]$} & $0.375${\scriptsize$[.35,.40]$} & $0.533${\scriptsize$[.51,.56]$} & $0.567${\scriptsize$[.54,.59]$} \\
10\%  & $0.450${\scriptsize$[.42,.48]$} & $0.481${\scriptsize$[.45,.51]$} & $0.554${\scriptsize$[.53,.58]$} & $0.593${\scriptsize$[.57,.62]$} \\
25\%  & $0.507${\scriptsize$[.48,.53]$} & $0.539${\scriptsize$[.51,.57]$} & $0.561${\scriptsize$[.54,.59]$} & $0.594${\scriptsize$[.57,.62]$} \\
50\%  & $0.505${\scriptsize$[.48,.53]$} & $0.553${\scriptsize$[.53,.58]$} & $0.558${\scriptsize$[.53,.59]$} & $0.601${\scriptsize$[.57,.63]$} \\
\midrule
\multicolumn{1}{l}{large-minus-tiny $\Delta$} & \multicolumn{2}{c}{${+}0.157$ vs.\ ${+}0.178$} & \multicolumn{2}{c}{${+}0.025$ vs.\ ${+}0.034$} \\
\bottomrule
\end{tabular}
\caption{\textbf{Unprefixed vs.\ prefixed catalog} (end-to-end, \texttt{raw}, \texttt{gpt-5.4-mini},
$N{=}1{,}279$): EX$(K)$ with bootstrap $95\%$ CIs over questions. Removing the \texttt{<db>} signal
lowers EX by a small uniform margin but preserves the budget-dependent pattern: lexical EX increases
with $K$, and its large-minus-tiny $95\%$ confidence interval excludes zero; dense EX remains near
saturation.}\label{tab:unprefixed}
\end{table}

\end{document}